\documentclass[conference]{IEEEtran}
\IEEEoverridecommandlockouts

\usepackage[T1]{fontenc}
\usepackage{cite}
\usepackage{amsmath,amssymb,amsfonts}
\usepackage{amsthm}
\usepackage{graphicx}
\usepackage{textcomp}
\usepackage{xcolor}

\newtheorem{lemma}{Lemma} 

\usepackage[ruled,vlined,linesnumbered]{algorithm2e}

\usepackage{subcaption} 
\usepackage{booktabs}
\usepackage{multirow}
\usepackage{rotating}
\usepackage{adjustbox}
\usepackage{makecell}
\setcellgapes{3pt}

\usepackage[colorlinks=true,allcolors=blue]{hyperref}
\usepackage{url}

\begin{document}

\title{Reversible Residual Normalization Alleviates \\ Spatio-Temporal Distribution Shift}

\author{
\IEEEauthorblockN{Zhaobo HU}
\IEEEauthorblockA{\textit{SAMOVAR, T\'el\'ecom SudParis} \\
\textit{Institut Polytechnique de Paris}\\
zhaobo.hu@telecom-sudparis.eu}
\and
\IEEEauthorblockN{Vincent Gauthier}
\IEEEauthorblockA{\textit{SAMOVAR, T\'el\'ecom SudParis} \\
\textit{Institut Polytechnique de Paris}\\
vincent.gauthier@telecom-sudparis.eu}
\and
\IEEEauthorblockN{Mehdi Naima}
\IEEEauthorblockA{\textit{CNRS -- LIP6} \\
\textit{Sorbonne Universit\'e}\\
mehdi.naima@lip6.com}
}


\maketitle

\begin{abstract}
Distribution shift severely degrades the performance of deep forecasting models. While this issue is well-studied for individual time series, it remains a significant challenge in the spatio-temporal domain. Effective solutions like instance normalization and its variants can mitigate temporal shifts by standardizing statistics. However, distribution shift on a graph is far more complex, involving not only the drift of individual node series but also heterogeneity across the spatial network where different nodes exhibit distinct statistical properties. To tackle this problem, we propose Reversible Residual Normalization (RRN), a novel framework that performs spatially-aware invertible transformations to address distribution shift in both spatial and temporal dimensions. Our approach integrates graph convolutional operations within invertible residual blocks, enabling adaptive normalization that respects the underlying graph structure while maintaining reversibility. By combining Bounded Scale Normalization with spectral-constrained graph neural networks, our method captures and normalizes complex Spatio-Temporal relationships in a data-driven manner. The bidirectional nature of our framework allows models to learn in a normalized latent space and recover original distributional properties through inverse transformation, offering a practical and model-agnostic solution for forecasting on dynamic spatio-temporal systems. 
\end{abstract}


\section{Introduction}
Spatio-temporal forecasting is fundamental to managing dynamic systems, ranging from urban traffic \cite{guo2019attention, shi2020spatial, fang2019gstnet} and environmental monitoring \cite{liang2023airformer, hu2022air, jin2023spatio} to public health surveillance. Effective models must capture both the spatial dependencies among interconnected locations and the temporal dynamics governing system evolution. Despite deep learning advancements, a critical challenge persists: distribution shift. This phenomenon occurs when statistical properties change, violating the assumption that training and testing data follow identical distributions. In spatio-temporal systems, this shift is dual-natured. Temporally, observations drift due to evolving dynamics; spatially, locations exhibit statistical heterogeneity despite sharing underlying correlations. This Spatio-Temporal distribution shift \cite{HuFYWXFW23} severely degrades generalization when networks trained on historical data confront evolved patterns during inference.

Current mitigation strategies primarily focus on time series, using methods like instance normalization \cite{ulyanov2016instance, kim2021reversible} to standardize sequences and remove non-stationary components. However, a critical gap emerges when these temporal techniques are directly transplanted to the spatio-temporal domain. Applying normalization independently to each node neglects spatial heterogeneity, where distinct geographic locations inherently possess different variance scales and implicitly disrupts the underlying topological dependencies of the graph. Simply put, standardizing a traffic hub and a suburban road with the same unconstrained temporal variance metrics erases their crucial spatial distinctions, confusing the downstream forecasting models.

To bridge this gap, we propose \textbf{Reversible Residual Normalization} (RRN), a novel framework that strictly extends normalizing flows to the spatio-temporal domain without sacrificing spatial correlations. The core innovation of RRN lies in replacing conventional normalizations with \textbf{Bounded Scale Normalization} (BSN). Unlike Instance Normalization, which introduces gradient singularities and violates invertibility conditions when temporal variance approaches zero, BSN adaptively stabilizes variance shifts while explicitly bounding the Lipschitz constant. We embed this module alongside spectral-constrained graph convolutions within invertible residual blocks. This bidirectional framework maps heterogeneous data into a partially stabilized latent space for forecasting, and reconstructs the spatial nuances through inverse transformation.

In summary, our main contributions are threefold:
\begin{itemize}
    \item We propose Bounded Scale Normalization (BSN) to adaptively scale temporal variance while mathematically guaranteeing Lipschitz continuity, resolving the instability of standard instance normalization in invertible architectures.
    \item We design Reversible Residual Normalization (RRN), an invertible framework that combines BSN and spectral-constrained graph convolutions. It explicitly addresses spatio-temporal distribution shifts while preserving the underlying graph topology.
    \item We evaluate RRN across multiple datasets and backbone architectures. The results demonstrate that our framework consistently improves forecasting performance as an effective plug-and-play module.
\end{itemize}

\section{Related Works}
\subsection{Distribution Shift in Time Series and Spatio-Temporal Data} 
Distribution shift arises from non-stationarity, causing significant discrepancies between training and testing distributions \cite{kim2021reversible, liu2023adaptive, fan2023dish, fan2025flow}. To mitigate this, instance normalization and its variants, such as reversible and adaptive normalization, effectively standardize temporal sequences by removing non-stationary components, yet they typically process nodes independently. Other advanced approaches explicitly model distribution shifts by learning intra-space and inter-space variations, but these remain confined to the temporal dimension. Consequently, the spatial dimension of distribution shift is largely neglected. Applying temporal normalization strategies directly to spatio-temporal data ignores the underlying graph structure, failing to account for spatial heterogeneity where distinct nodes exhibit unique statistical properties.

\subsection{Invertible Networks}
Invertible neural networks learn bijective transformations that allow for exact computation of both forward and inverse mappings, facilitating tasks such as generative modeling and density estimation \cite{dinh2017density, zhai2025normalizing, liu2019graph, behrmann2019invertible, kingma2018glow}. While normalizing flows typically rely on coupling layers to construct invertible mappings \cite{dinh2017density, kingma2018glow, nice_2015ICLR, kingma2016improved}, Invertible Residual Networks \cite{behrmann2019invertible, zha2021invertible, park2024mitigating} extend this capability to standard residual architectures. Rather than relying on partitioning dimensions, these models utilize residual blocks \cite{he2016residual} that guarantee invertibility provided the residual function satisfies specific Lipschitz continuity conditions (requiring a constant strictly less than one). This constraint is often enforced via spectral normalization to enable exact inverse computation through fixed-point iteration. Despite their success in time series, the potential of invertible architectures to simultaneously address spatial and temporal distribution shift while preserving graph topology remains underexplored.

\section{Problem Formulations}
\subsection{Spatio-Temporal Forecasting}
Let a spatio-temporal system be represented by a graph $\mathcal{G} = (V, E)$, where $V$ is a set of $N$ spatial locations (e.g., sensors, regions) with $|V|=N$, and $E$ is a set of edges representing the spatial relationships between these locations. At each discrete time step $t$, the system exhibits a feature matrix $X^{(t)} \in \mathbb{R}^{N \times D}$, where $D$ is the number of features recorded at each location. This feature matrix is also referred to as a graph signal.

The task of spatio-temporal forecasting aims to predict future system states based on historical observations. Given a lookback window of length $L$, the historical data can be denoted as a tensor $\mathcal{X}_{t-L+1:t} = (X^{(t-L+1)}, \dots, X^{(t)}) \in \mathbb{R}^{L \times N \times D}$. The objective is to forecast the subsequent data over a horizon window of length $H$, denoted as $\mathcal{Y}_{t+1:t+H} = (X^{(t+1)}, \dots, X^{(t+H)}) \in \mathbb{R}^{H \times N \times D}$.

Formally, the goal is to learn a function $f_{\theta}$ parameterized by $\theta$ that maps the historical observations and the graph structure to the future sequence:
\begin{equation*}
    \hat{\mathcal{Y}}_{t+1:t+H} = f_{\theta}(\mathcal{X}_{t-L+1:t}; \mathcal{G})
    \label{eq:st_forecasting}
\end{equation*}
where $\hat{\mathcal{Y}}_{t+1:t+H}$ represents the predicted future sequence.

\subsection{Distribution Shift in Spatio-Temporal Series}
A primary challenge in real-world spatio-temporal forecasting is the non-stationarity of the data, which manifests as a distribution shift. Many deep learning models implicitly assume that the data-generating process is consistent between the training and test periods, an assumption that is frequently violated in practice. This discrepancy occurs as the joint distribution of the observed spatio-temporal data changes over time, hindering a model's ability to generalize from past observations to future predictions.

Formally, for any two distinct time steps $t_u$ and $t_v$, the underlying conditional probability that governs the system's evolution is not constant. This distribution shift can be expressed as:
$$
P({\mathcal{Y}}_{t_u+1:t_u+H} | {\mathcal{X}}_{t_u-L+1:t_u}; \mathcal{G}) \neq P({\mathcal{Y}}_{t_v+1:t_v+H} | {\mathcal{X}}_{t_v-L+1:t_v}; \mathcal{G})
$$
Therefore, our objective is not only to predict future spatio-temporal states but also to explicitly model and mitigate these distribution shifts to enhance forecasting accuracy and generalization. To address this, we propose extending the principles of normalizing flows, which have shown promise in handling temporal distribution shifts, to the more complex spatio-temporal domain.

\section{Methodology}
\subsection{Invertible Residual Structure}

\begin{figure*}[t]
    \centering
    \includegraphics[width=\textwidth,keepaspectratio]{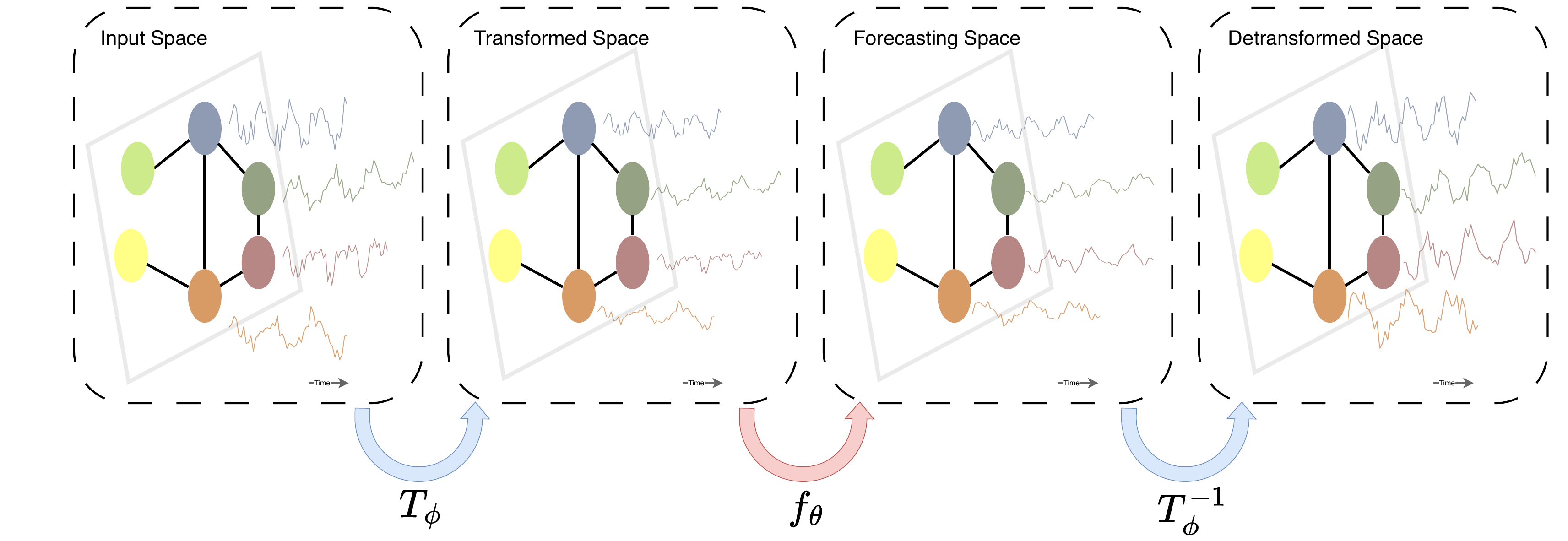}
    \caption{Overview of the Reversible Residual Normalization framework for spatio-temporal forecasting. The framework operates through four stages: (1) \textbf{Input Space} contains raw spatio-temporal data with distribution shifts across nodes and time; (2) \textbf{Transformed Space} where the invertible transformation $T_\phi$ normalizes the data by removing Spatio-Temporal distribution shifts while preserving correlational structures; (3) \textbf{Forecasting Space} where any forecasting model $f_\theta$ makes predictions in the stationary latent space; (4) \textbf{Detransformed Space} where the inverse transformation $T_\phi^{-1}$ restores the original distributional properties to produce final predictions.}
    \label{fig:Structure}
\end{figure*}

Residual structures \cite{he2016residual} have become a fundamental building block in modern deep learning architectures. A residual block takes the form:
\begin{equation}
    H(x) = x + G(x)
\end{equation}
Without loss of generality, we define $x \in \mathbb{R}^d$ is the input, $G: \mathbb{R}^d \to \mathbb{R}^d$ represents a learnable transformation, and the output $H(x)$ combines the identity mapping with the learned residual $G(x)$. This formulation facilitates gradient flow during training and enables the construction of very deep networks.

For our purposes, we are particularly interested in when such residual structures are invertible, meaning there exists a unique inverse mapping $H^{-1}$ such that $H^{-1}(H(x)) = x$ for all $x$ in the domain. Invertibility is desirable in our context because it allows us to transform data distributions bidirectionally, which is essential for modeling distribution shifts in spatio-temporal forecasting.

A sufficient condition for the invertibility of a residual block is provided by the following lemma, which constrains the behavior of the residual function $G$:

\begin{lemma}[Sufficient Condition for Invertible Residual Blocks \cite{behrmann2019invertible, park2024mitigating}]
\label{lemma:invertible_residual}
Let $H(x) = x + G(x)$ be a residual block, where $G: \mathbb{R}^d \to \mathbb{R}^d$. If the Lipschitz constant of $G$ satisfies:
\begin{equation}
    L(G) = \sup_{x_1 \neq x_2} \frac{\|G(x_1) - G(x_2)\|}{\|x_1 - x_2\|} < 1
\end{equation}
then $H$ is invertible.
\end{lemma}

The intuition behind this lemma is that when $G$ is a contraction mapping (i.e., $L(G) < 1$), the residual block $H$ becomes a perturbation of the identity function that preserves bijectivity. Under this condition, the inverse $x = H^{-1}(z)$ for a given output $z$ can be computed through fixed-point iteration as shown in Algorithm \ref{alg:inverse_residual}.

\begin{algorithm}
\caption{Inverse of Residual Block via Fixed-Point Iteration}
\label{alg:inverse_residual}
\KwIn{Output $\mathbf{x}^{(\ell)}$ from residual block, residual function $G$, number of iterations $N$}
\KwOut{input of residual block $\mathbf{x}^{(\ell-1)}$}
$\mathbf{x} \leftarrow \mathbf{x}^{(\ell)}$\;
\For{$m = 1, \ldots, N$}{
    $\mathbf{x} \leftarrow \mathbf{x} - G(\mathbf{x})$\;
}
\Return{$\mathbf{x}$}
\end{algorithm}

In practice, enforcing the Lipschitz constraint $L(G) < 1$ requires careful design of the residual function $G$. When $G$ is composed of linear transformations (such as convolutions) followed by nonlinear activations, a common approach is to constrain the spectral norm of each linear layer. Specifically, if $G$ consists of multiple layers with weight matrices $W_1, W_2, \ldots, W_i$, we can enforce:

\begin{equation}
\label{equa:spectral_norm}
\tilde{W}_i = 
\begin{cases}
c W_i / \tilde{\sigma}_i, & \text{if } c/\tilde{\sigma}_i < 1 \\
W_i, & \text{else}
\end{cases}
\end{equation}

where $\tilde{\sigma}_i=\|W_i\|_2$ denotes the spectral norm (largest singular value) of $W_i$, and $c < 1$ are chosen such that the overall Lipschitz constant of $G$ remains below 1. The spectral norm can be efficiently approximated using power iteration methods \cite{MiyatoKKY18, s10994-020-05929-w} during training. This framework of invertible residual blocks provides the foundation for our approach to handling distribution shifts in spatio-temporal forecasting, as we will elaborate in the following sections.

\subsection{Bounded Scale Normalization for Invertibility}

Instance Normalization (IN) \cite{ulyanov2016instance, fan2025flow, liu2023adaptive, kim2021reversible} has demonstrated significant effectiveness in mitigating distribution shifts by standardizing features across the temporal dimension. However, standard IN is inherently incompatible with invertible residual networks due to its violation of Lipschitz continuity (Lemma~\ref{lemma:invertible_residual}).

For input $\mathcal{X}_{t-T+1:t} \in \mathbb{R}^{T \times N \times D}$ representing temporal features at spatial locations, we omit the time subscripts for brevity and denote it as $\mathcal{X}$. Instance Normalization centers the data $y = \left(I - \frac{1}{T}\mathbf{1}\mathbf{1}^\top\right)\mathcal{X}$ and scales it by its standard deviation: $z = y / \text{Std}(y)$. The Jacobian matrix of this transformation reveals a critical instability:
\begin{equation}
J_z(\mathcal{X}) = \frac{\partial z}{\partial \mathcal{X}} = \frac{1}{\text{Std}(y)} \left(I - \frac{yy^\top}{\|y\|_2^2}\right) \left(I - \frac{1}{T}\mathbf{1}\mathbf{1}^\top\right).
\end{equation}
When the temporal sequence is highly stable (i.e., $\text{Std}(y) \to 0$), the entries of the Jacobian matrix approach infinity. This unboundedness breaks the Lipschitz condition ($\text{Lip} < 1$), causing the fixed-point iteration in the inverse transformation to diverge.

To overcome this instability while still addressing variance shifts, we propose \textbf{Bounded Scale Normalization} (BSN). BSN introduces a simple yet effective variance scaling mechanism that mathematically guarantees Lipschitz continuity:
\begin{equation}
\text{BSN}(\mathcal{X}) = \gamma \odot \left( \alpha \cdot \delta \frac{y}{\text{softplus}(\text{Std}(y) - \delta) + \delta} \right) + \beta,
\end{equation}
where $y$ is the mean-centered input, $\delta > 0$ is a predefined or learnable threshold that acts as a strict lower bound for the denominator, $\alpha$ is a controllable scaling hyperparameter, and $\gamma, \beta \in \mathbb{R}^{N \times D}$ are learnable affine parameters.

The core advantage of BSN lies in its bounded scaling factor. Because the function $\text{softplus}(\cdot)$ is strictly positive, the denominator $D = \text{softplus}(\text{Std}(y) - \delta) + \delta$ satisfies $D > \delta$ universally. Consequently, the scaling factor applied to the centered data $y$ is strictly bounded by $\frac{\alpha \delta}{\delta} = \alpha$. Assuming $\gamma = \mathbf{1}$ and $\beta = \mathbf{0}$ for the base transformation, the Lipschitz constant of the BSN module is strictly constrained:
\begin{equation}
\text{Lip}(\text{BSN}) \le \alpha.
\end{equation}

By setting $\alpha < 1$ (e.g., $\alpha = 0.9$), BSN satisfies the necessary contraction mapping condition for invertibility in Lemma~\ref{lemma:invertible_residual}. In practice, when the variance is large, BSN behaves proportionally to standard normalization, as the constant scaling factor $\alpha \delta$ is easily absorbed by the learnable parameter $\gamma$. Conversely, when the variance approaches zero, the denominator is safely floored by $\delta$. This fundamentally prevents the division-by-zero issue, ensures stable gradient flow during training, and preserves IN's ability to mitigate temporal distribution shifts.

\subsection{Overall Structure}

After applying Bounded Scale Normalization, we incorporate a graph convolution module within the invertible residual block to capture spatial dependencies. Following the framework in Section 3.1, we introduce a Lipschitz-constrained Graph Convolutional Network (GCN) layer \cite{Kipf:2016tc, park2024mitigating} to ensure the overall residual block satisfies the invertibility condition.

\subsubsection{Spectral Normalization for GCN}

A standard GCN layer with residual connection can be formulated as:
\begin{equation}
\begin{split}
H(X^{(t)}) &= X^{(t)} + \sigma(\hat{A}X^{(t)}W) \\
           &= X^{(t)} + g(X^{(t)}),
\end{split}
\label{eq:gcn_residual}
\end{equation}
where $X^{(t)} \in \mathbb{R}^{N \times D}$ is the input node representation, we omit the time superscripts for brevity and denote it as $X$, $\hat{A} = \tilde{D}^{-\frac{1}{2}}\tilde{A}\tilde{D}^{-\frac{1}{2}}$ is the normalized adjacency matrix with $\tilde{A} = A + I$ (adjacency matrix with self-loops) and $\tilde{D}$ is the diagonal degree matrix, $W \in \mathbb{R}^{d \times d}$ is the learnable weight matrix, and $\sigma(\cdot)$ is a Lipschitz continuous activation function (e.g., ReLU, tanh). According to Lemma~\ref{lemma:invertible_residual}, the residual block in Eq.~\eqref{eq:gcn_residual} is invertible if the Lipschitz constant of the residual function $g(X) = \sigma(\hat{A}XW)$ satisfies:
\begin{equation}
\text{Lip}(g) = \sup_{X_1 \neq X_2} \frac{\|g(X_1) - g(X_2)\|_2}{\|X_1 - X_2\|_2} < 1.
\label{eq:lip_condition}
\end{equation}

For contractive activation functions like ReLU and tanh where $\text{Lip}(\sigma) <  1$, the condition is satisfied if:
\begin{equation}
\sup_{X \neq 0} \frac{\|\hat{A}XW\|_2}{\|X\|_2} < 1.
\label{eq:lip_gcn}
\end{equation}

This supremum is upper bounded by:
\begin{equation}
\sup_{X \neq 0} \frac{\|\hat{A}XW\|_2}{\|X\|_2} \leq \|\hat{A}\|_2 \|W\|_F,
\label{eq:upper_bound}
\end{equation}

where $\|\cdot\|_2$ denotes the spectral norm (the largest singular value), and $\|W\|_F$ is the Frobenius norm of the weight matrix. While strict invertibility only requires the spectral norm to be bounded ($\|W\|_2 < 1$), computing the exact spectral norm during each forward pass via power iteration introduces computational overhead, especially for deep spatio-temporal graphs. To achieve computational efficiency without relaxing the theoretical guarantee, we enforce a constraint on the Frobenius norm instead. Mathematically, the Frobenius norm acts as a strict upper bound for the spectral norm:
\begin{equation}
|W|_2 \le |W|_F
\end{equation}
By normalizing the weights such that $\|W\|_F \le c$ (with $c < 1$), we universally guarantee that $\|W\|_2 < 1$. This design choice acts as a more stringent regularization. It bypasses the iterative estimation of singular values, accelerating training throughput \cite{park2024mitigating}, while satisfying the contraction mapping condition required by Lemma \ref{lemma:invertible_residual}.

\begin{figure}[htbp]
    \centering
    \includegraphics[width=1.0\columnwidth]{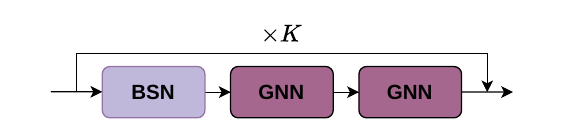} 
    \caption{Invertible Residual Block architecture.}
    \label{fig:Residual_block}
\end{figure}

\subsubsection{Reversible Residual Normalization}

We now combine Bounded Scale Normalization (BSN) and the Lipschitz-constrained GCN into an invertible residual block. The complete block is defined as:
\begin{equation}
H(\mathcal{X}_{t-T+1:t}^{(\ell)}) = \mathcal{X}_{t-T+1:t}^{(\ell)} + \sigma(\hat{A} \cdot \text{BSN}(\mathcal{X}_{t-T+1:t}^{(\ell)}) \cdot W),
\label{eq:invertible_st_block}
\end{equation}
where $\text{BSN}(\cdot)$ is the Bounded Scale Normalization introduced previously. According to the chain rule for Lipschitz continuity, the Lipschitz constant of the residual function $g(\mathcal{X}) = \sigma(\hat{A} \cdot \text{BSN}(\mathcal{X}) \cdot W)$ satisfies:
\begin{equation}
\begin{split}
\text{Lip}(g) &\leq \text{Lip}(\sigma) \cdot \|\hat{A}\|_2 \cdot \text{Lip}(\text{BSN}) \cdot \|W\|_F \\
&\leq \text{Lip}(\sigma) \cdot 1 \cdot \alpha \cdot \|W\|_F.
\end{split}
\label{eq:block_lipschitz}
\end{equation}
Since $\text{Lip}(\sigma) \leq 1$ for standard contractive activations (e.g., ReLU), the condition $\text{Lip}(g) < 1$ required for invertibility is strictly guaranteed as long as we constrain the weight matrix such that $\alpha \cdot \|W\|_F < 1$.

\subsection{Bidirectional Transformation Framework}

Since our framework operates without global static preprocessing (e.g., z-score scaling), the identity mapping in the residual block (Eq.~\eqref{eq:invertible_st_block}) inherently preserves the original raw magnitudes of the input. Consequently, unscaled extreme values can bypass the bounded residual branch, propagating through the deep network and destabilizing gradients during training. 

To suppress these extreme values while maintaining exact reversibility, we introduce an initial Activation Normalization (ActNorm) \cite{kingma2018glow} module prior to the deep residual blocks. It dynamically standardizes inputs using instance-specific statistics purely to squash extreme magnitudes into a stable computational range. Unlike approaches that rely on instance normalization as the primary mechanism for distribution shift, ActNorm serves strictly as a numerical stabilizer for the Lipschitz-constrained network. Because it tracks these scaling statistics, its exact inverse losslessly restores the original physical scales at the end of the pipeline.

Our complete bidirectional framework therefore transforms raw spatio-temporal data into a stationary latent space, performs forecasting, and transforms the predictions back to the original space. Formally, let $\mathcal{X}_{t-L+1:t}$ denote the historical observations:

\textbf{Forward Transformation (to stationary space):}
\begin{equation}
\mathcal{Z} = T_\phi(\mathcal{X}_{t-L+1:t}; \mathcal{G}),
\label{eq:forward_transform}
\end{equation}
where $T_\phi$ is composed of the initial ActNorm followed by $M$ invertible spatio-temporal blocks: $T_\phi = H^{(M)} \circ \dots \circ H^{(1)} \circ \text{ActNorm}$, with $H^{(\ell)}$ defined in Eq.~\eqref{eq:invertible_st_block}.\\

\textbf{Prediction in stationary space:}
\begin{equation}
\hat{\mathcal{Z}}_{t+1:t+H} = f_{\theta}(\mathcal{Z}; \mathcal{G}),
\label{eq:stationary_prediction}
\end{equation}
where $f_{\theta}$ is an arbitrary spatio-temporal forecasting model operating in the stabilized latent space.\\

\textbf{Inverse Transformation (back to original space):}
\begin{equation}
\hat{\mathcal{Y}}_{t+1:t+H} = T_\phi^{-1}(\hat{\mathcal{Z}}_{t+1:t+H}; \mathcal{G}),
\label{eq:inverse_transform}
\end{equation}
where the inverse mapping is $T_\phi^{-1} = \text{ActNorm}^{-1} \circ (H^{(1)})^{-1} \circ \dots \circ (H^{(M)})^{-1}$. Each residual inverse $(H^{(\ell)})^{-1}$ is computed via fixed-point iteration to recover the complex spatio-temporal relationships, and the final $\text{ActNorm}^{-1}$ step explicitly scales the predictions back to restore the true physical magnitudes.

The key insight is that by ensuring numerical stability via ActNorm and subsequently transforming the data through our bounded spatio-temporal residual blocks, we mitigate the complex spatio-temporal distribution shift. The latent conditional distribution $P(\hat{\mathcal{Z}}_{t+1:t+H} | \mathcal{Z}; \mathcal{G})$ becomes robust and approximately invariant, enabling accurate forecasting with any backbone model $f_{\theta}$.

\section{Experiments}

\subsection{Experimental Setup}
\noindent\textbf{Datasets}\quad We evaluate our framework on four diverse spatio-temporal benchmarks. \textbf{METR-LA} and \textbf{PEMS-BAY} are classic traffic speed datasets containing 207 and 325 sensors from Los Angeles and the San Francisco Bay Area, respectively, with 5-minute sampling frequency over 4--6 months. \textbf{SDWPF} \cite{zhou2024sdwpf} represents an energy domain, providing 24 months of wind power data from 134 turbines, featuring 19 dynamic variables including SCADA system parameters and ERA5 meteorological data. \textbf{LargeST-SD} \cite{liu2023largest} is a large-scale traffic flow dataset encompassing 716 sensors in San Diego over a 5-year horizon (2017--2021) aggregated at 15-minute intervals. Regarding graph construction, METR-LA, PEMS-BAY, and LargeST-SD utilize distance-based connectivity via Gaussian kernel thresholding, while SDWPF leverages spatial proximity between turbines. For data splitting, METR-LA, PEMS-BAY, and SDWPF are chronologically split into 70\% training, 10\% validation, and 20\% testing sets. LargeST-SD is split by years, utilizing the full year of 2019 for training, alongside 3-month blocks of 2020 and 2021 for validation and testing, respectively.\\

\noindent\textbf{Baselines}\quad We evaluate our model against several baseline methods addressing distribution shift. \textbf{RevIN}~\cite{kim2021reversible} normalizes each instance by its mean and variance, symmetrically denormalizing outputs to restore original scales. \textbf{Dish-TS}~\cite{fan2023dish} models intra-space and inter-space shifts using a Dual-CONET framework to learn separate distribution coefficients. \textbf{SAN}~\cite{liu2023adaptive} adopts a fine-grained approach, normalizing sub-series "slices" while predicting future statistics for adaptive denormalization. \textbf{ST-Norm}~\cite{deng2021st} utilizes temporal and spatial modules to separately refine high-frequency and local components by removing low-frequency and global trends from raw data. \textbf{IN-Flow}~\cite{fan2025flow} employs an instance normalization flow within a decoupled bi-level optimization framework to reversibly transform non-stationary time series distributions. \textbf{SAUP}~\cite{HuFYWXFW23} explicitly mitigates spatial-temporal shifts by utilizing invertible attention flows to project data into a unified stationary space before capturing topological and geographic correlations.\\

\noindent\textbf{Backbone models}\quad To demonstrate the versatility of our model-agnostic framework, we evaluate it on diverse mainstream architectures. \textbf{DCRNN}~\cite{li2018dcrnn_traffic} integrates diffusion convolution with GRUs to model traffic as a directed diffusion process. \textbf{Graph WaveNet}~\cite{wu2019graph} combines dilated TCNs with an adaptive adjacency matrix to capture hidden spatial dependencies. \textbf{GMAN}~\cite{zheng2020gman} utilizes an encoder-decoder structure with spatio-temporal attention blocks to model dynamic correlations. \textbf{Auto-DSTSGN}~\cite{jin2022automated} employs an automated dilated graph framework with neural architecture search to construct adaptive adjacency matrices. \textbf{STPGNN}~\cite{kong2024spatio} introduces a pivotal graph convolution module to prioritize sensors with complex dependencies within a parallel feature extraction framework.\\

\noindent\textbf{Implementation Details}\quad We follow the original configurations for all backbone models. For RRN, we stack two invertible residual blocks with a hidden size of 32. To guarantee invertibility (Lemma~\ref{lemma:invertible_residual}), we set the Bounded Scale Normalization (BSN) scaling $\alpha=0.9$ and constrain the GCN weights to a Frobenius norm of 0.9. Models are trained using the Adam optimizer with a learning rate of $5 \times 10^{-3}$ and a batch size of 64 for up to 100 epochs with early stopping. Experiments are conducted on NVIDIA L40S GPUs, and we report the average performance over five random seeds.

\subsection{Main Results}

\begin{table*}[htbp]
\centering
\setlength{\belowcaptionskip}{10pt}
\caption{Forecasting performance comparison across different datasets and models. All results are the average of 5 random seeds. Bold values indicate better performance in each baseline vs. RRN pair (lower is better).}
\label{tab:results_updated}
\renewcommand{\arraystretch}{2.0}
\begin{adjustbox}{max width=\textwidth}
\begin{tabular}{@{}cccccccccccccccccccccc@{}}
\toprule
\multirow{2}{*}{\textbf{Dataset}} & \multirow{2}{*}{\textbf{Horizon}} & \multicolumn{2}{c}{\textbf{GWavenet}} & \multicolumn{2}{c}{\textbf{GWavenet + RRN}} & \multicolumn{2}{c}{\textbf{DCRNN}} & \multicolumn{2}{c}{\textbf{DCRNN + RRN}} & \multicolumn{2}{c}{\textbf{Auto-DSTSGN}} & \multicolumn{2}{c}{\textbf{Auto-DSTSGN+RRN}} & \multicolumn{2}{c}{\textbf{GMAN}} & \multicolumn{2}{c}{\textbf{GMAN + RRN}} & \multicolumn{2}{c}{\textbf{STPGNN}} & \multicolumn{2}{c}{\textbf{STPGNN + RRN}} \\
\cmidrule(lr){3-4} \cmidrule(lr){5-6} \cmidrule(lr){7-8} \cmidrule(lr){9-10} \cmidrule(lr){11-12} \cmidrule(lr){13-14} \cmidrule(lr){15-16} \cmidrule(lr){17-18} \cmidrule(lr){19-20} \cmidrule(lr){21-22}
& & MAE & RMSE & MAE & RMSE & MAE & RMSE & MAE & RMSE & MAE & RMSE & MAE & RMSE & MAE & RMSE & MAE & RMSE & MAE & RMSE & MAE & RMSE \\
\midrule
\multirow{3}{*}{\rotatebox[origin=c]{90}{\textbf{SDWPF}}}
& 3  & 56.03 & 107.19 & \textbf{54.86} & \textbf{104.88} & 62.52 & 120.13 & \textbf{60.89} & \textbf{117.85} & 56.62 & 105.91 & \textbf{55.26} & \textbf{104.02} & 58.08 & 108.58 & \textbf{55.84} & \textbf{106.36} & 57.67 & 109.48 & \textbf{56.42} & \textbf{108.30} \\
& 6  & 82.14 & 147.08 & \textbf{79.94} & \textbf{146.30} & 89.83 & 161.09 & \textbf{87.41} & \textbf{160.92} & 80.52 & 145.92 & \textbf{78.91} & \textbf{144.81} & 80.90 & 147.55 & \textbf{80.70} & \textbf{147.50} & 81.38 & 149.70 & \textbf{80.65} & \textbf{146.70} \\
& 12 & 117.24 & 202.14 & \textbf{116.08} & \textbf{199.41} & 126.81 & 216.88 & \textbf{124.96} & \textbf{212.48} & 113.71 & 198.42 & \textbf{113.32} & \textbf{195.55} & 116.55 & 200.50 & \textbf{115.78} & \textbf{199.83} & 116.57 & 203.49 & \textbf{116.39} & \textbf{199.40} \\
\midrule
\multirow{3}{*}{\rotatebox[origin=c]{90}{\textbf{LargeST-SD}}}
& 3  & 19.60 & 32.19 & \textbf{17.28} & \textbf{28.31} & 20.55 & 33.26 & \textbf{19.93} & \textbf{31.46}  & 19.39 & 31.42 & \textbf{18.66} & \textbf{29.62} & OOM & OOM & OOM & OOM & 20.27 & 32.31 & \textbf{20.04} & \textbf{31.91}  \\
& 6  & 28.30 & 47.16 & \textbf{24.14} & \textbf{39.55}  & 30.50 & 48.45 & \textbf{28.66} & \textbf{44.75}  & 25.95 & 41.37 & \textbf{25.42} & \textbf{39.76} & OOM & OOM & OOM & OOM & 29.04 & 45.57 & \textbf{27.94} & \textbf{44.02}  \\
& 12 & 43.35 & 70.07 & \textbf{36.69} & \textbf{56.99}  & 48.70 & 71.59 & \textbf{46.26} & \textbf{69.66} & 38.89 & 56.75 & \textbf{36.65} & \textbf{56.53} & OOM & OOM & OOM & OOM & 41.11  & 61.53 & \textbf{40.62} & \textbf{61.25} \\
\midrule
\multirow{3}{*}{\rotatebox[origin=c]{90}{\textbf{MetrLA}}}
& 3  & 2.85 & 5.42 & \textbf{2.76} & \textbf{5.26}  & 2.90 & 5.58 & \textbf{2.89} & \textbf{5.47} & 2.83 & 5.39 & \textbf{2.78} & \textbf{5.38} & 2.81 & 5.46 & \textbf{2.76} & \textbf{5.32} & 3.04 & 5.73 & \textbf{3.00} & \textbf{5.70} \\
& 6  & 3.20 & 6.38 & \textbf{3.16} & \textbf{6.34}  & 3.39 & 6.70 & \textbf{3.36} & \textbf{6.57} & 3.19 & 6.39 & \textbf{3.15} & \textbf{6.35} & 3.19 & 6.48 & \textbf{3.15} & \textbf{6.40} & 3.49 & 6.83 & \textbf{3.43} & \textbf{6.83} \\
& 12 & 3.69 & 7.49 & \textbf{3.64} & \textbf{7.42} & 4.09 & 8.10 & \textbf{4.04} & \textbf{7.96} & 3.66 & 7.44 & \textbf{3.57} & \textbf{7.37} & 3.63 & 7.61 & \textbf{3.61} & \textbf{7.50}  & 4.06 & 8.13 & \textbf{3.96} & \textbf{7.92} \\
\midrule
\multirow{3}{*}{\rotatebox[origin=c]{90}{\textbf{PemsBay}}}
& 3  & 1.32 & 2.79 & \textbf{1.32} & \textbf{2.77} & 1.41 & 2.94 & \textbf{1.40} & \textbf{2.94} & 1.38 & 2.88 & \textbf{1.32} & \textbf{2.80} & 1.36 & 2.88 & \textbf{1.34} & \textbf{2.82} & 1.50 & 3.08 & \textbf{1.39} & \textbf{2.93}  \\
& 6  & 1.65 & 3.73 & \textbf{1.64} & \textbf{3.67}  & 1.80 & 3.98 & \textbf{1.79} & \textbf{3.98}  & 1.71 & 3.79 & \textbf{1.67} & \textbf{3.71} & 1.70 & 3.84 & \textbf{1.67} & \textbf{3.67} & 1.89 & 4.07 & \textbf{1.77} & \textbf{3.93}  \\
& 12 &  1.98 & 4.49 & \textbf{1.95} & \textbf{4.40}  & 2.29 & 5.10 & \textbf{2.26} & \textbf{5.07} & 2.02 & 4.56 & \textbf{1.99} & \textbf{4.46} & 2.03 & 4.57 & \textbf{1.97} & \textbf{4.48} & 2.28 & 4.94 & \textbf{2.15} & \textbf{4.81}  \\
\bottomrule
\end{tabular}
\end{adjustbox}
\end{table*}

\begin{table*}[t]
\centering
\setlength{\belowcaptionskip}{10pt} 
\caption{Performance comparison of DCRNN with various enhancement methods across different datasets. All results represent the average of five random seeds (lower is better).}
\label{tab:dcrnn_results}
\renewcommand{\arraystretch}{2.0}
\begin{adjustbox}{max width=\textwidth}
\begin{tabular}{@{}cccccccccccccccccc@{}}
\toprule
\multirow{2}{*}{Dataset} & \multirow{2}{*}{Horizon} & \multicolumn{2}{c}{DCRNN} & \multicolumn{2}{c}{DCRNN + RevIN} & \multicolumn{2}{c}{DCRNN + Dish-TS} & \multicolumn{2}{c}{DCRNN + SAN} & \multicolumn{2}{c}{DCRNN + STNORM} & \multicolumn{2}{c}{DCRNN + IN-Flow} & \multicolumn{2}{c}{DCRNN + SAUP} & \multicolumn{2}{c}{DCRNN + RRN} \\
\cmidrule(lr){3-4} \cmidrule(lr){5-6} \cmidrule(lr){7-8} \cmidrule(lr){9-10} \cmidrule(lr){11-12} \cmidrule(lr){13-14} \cmidrule(lr){15-16} \cmidrule(lr){17-18}
& & MAE & RMSE & MAE & RMSE & MAE & RMSE & MAE & RMSE & MAE & RMSE & MAE & RMSE & MAE & RMSE & MAE & RMSE \\
\midrule

\multirow[c]{3}{*}{\rotatebox[origin=c]{90}{SDWPF}}
& 3  & 62.52 & 120.13          & 62.07  & 120.69 & 61.51  & 118.76 & 90.50  & 152.42 & 63.62   & 121.13 & 90.51 & 158.16 & 70.67  & 130.14 & \textbf{60.89} & \textbf{117.85} \\
& 6  & 89.83 & 161.09          & 88.98  & 164.55 & 87.90  & 161.18 & 115.74 & 190.34 & 89.91  & 162.84 & 104.27 & 221.09 & 92.81  & 167.41 & \textbf{87.41} & \textbf{160.92} \\
& 12 & 126.81& 216.88 & 127.88 & 221.80 & 125.19 & 215.95 & 143.55 & 231.31 & 125.54   & 214.98 & 152.73 & 227.43 & 126.55 & 215.82 & \textbf{124.96}&  \textbf{212.48}\\
\midrule

\multirow[c]{3}{*}{\rotatebox[origin=c]{90}{LargeST-SD}}
& 3  & 20.55 & 33.26 & 20.88 & 33.94 & 21.54 & 35.00 & 23.58 & 37.45 & 20.55 & 33.26 & 45.63 & 46.62 & OOM & OOM & \textbf{19.93} & \textbf{31.46} \\
& 6  & 30.50 & 48.45 & 33.06 & 52.99 & 31.92 & 52.21 & 35.08 & 53.64 & 30.50 & 48.45  & 57.26 & 66.69 & OOM & OOM & \textbf{28.66} & \textbf{44.75} \\
& 12 & 48.70 & 71.59 & 58.07 & 87.11 & 53.36 & 88.33 & 64.32 & 92.15 & 47.14 & 70.85   & 62.13 & 77.13 & OOM & OOM & \textbf{46.26} & \textbf{69.66} \\
\midrule

\multirow[c]{3}{*}{\rotatebox[origin=c]{90}{MetrLA}}
& 3  & 2.90 & 5.58 & 2.97 & 5.75 & 2.95 & 5.60 & 3.15 & 6.12 & 2.90 & 5.53 & 4.56 & 7.51 & 3.12 & 5.94 & \textbf{2.89} & \textbf{5.47} \\
& 6  & 3.39 & 6.70 & 3.57 & 7.13 & 3.48 & 6.82 & 3.83 & 7.63 & 3.39 & 6.67 & 6.77 & 9.24 & 3.76 & 7.43 & \textbf{3.36} & \textbf{6.57} \\
& 12 & 4.09 & 8.10 & 4.54 & 9.01 & 4.33 & 8.49 & 5.10 & 9.95 & 4.06 & 8.03 & 7.48 & 10.89 & 4.81 & 9.49 & \textbf{4.04} & \textbf{7.96} \\
\midrule

\multirow[c]{3}{*}{\rotatebox[origin=c]{90}{PemsBay}}
& 3  & 1.41 & 2.94 & 1.43 & 3.08 & 1.45 & 3.04 & 1.60 & 3.35 & 1.40 & 2.97 & 3.08 & 4.22 & OOM & OOM & \textbf{1.40} & \textbf{2.94} \\
& 6  & 1.79 & 4.01 & 1.91 & 4.39 & 1.88 & 4.21 & 2.12 & 4.72 & 1.80 & 3.99 & 3.75 & 5.16 & OOM & OOM & \textbf{1.79} & \textbf{3.98} \\
& 12 & 2.27 & 5.10 & 2.60 & 5.97 & 2.48 & 5.55 & 2.98 & 6.65 & 2.26 & 5.09 & 4.81 & 6.57 & OOM & OOM & \textbf{2.26} & \textbf{5.07} \\
\bottomrule
\end{tabular}
\end{adjustbox}
\end{table*}

The experimental results, summarized in Table~\ref{tab:results_updated} and Table~\ref{tab:dcrnn_results}, demonstrate the effectiveness of our Reversible Residual Normalization (RRN) framework in mitigating spatio-temporal distribution shifts.

\noindent\textbf{Performance Gains Across Backbones.}\quad As shown in Table~\ref{tab:results_updated}, RRN consistently enhances forecasting accuracy across all five backbone models and three datasets. The improvements are notable on the SDWPF dataset, which features high volatility due to meteorological changes; for instance, GWavenet's MAE decreases by approximately 7.7\% and 5.4\% for the 3-step and 6-step horizons, respectively. These consistent improvements across diffusion-based, attention-based, and automated-search architectures demonstrate RRN's utility as a model-agnostic plugin for stabilizing latent representations.

\noindent\textbf{Comparison with Normalization Strategies.}\quad We evaluate RRN against various normalization methods using DCRNN in Table~\ref{tab:dcrnn_results}. First, RRN outperforms other invertible approaches (RevIN, Dish-TS, SAN) by integrating graph-aware operations, whereas these baselines typically treat nodes as independent time series. Second, the non-invertible method \textbf{ST-Norm} shows competitive performance, occasionally matching RRN on the SDWPF dataset. ST-Norm's success in filtering high-frequency noise highlights the value of spatio-temporal decoupling, suggesting a promising direction for integrating multi-frequency decomposition into our invertible framework in future work.

\subsection{Detailed Analysis}

\begin{figure}[t]
    \centering
    \includegraphics[width=\columnwidth,keepaspectratio]{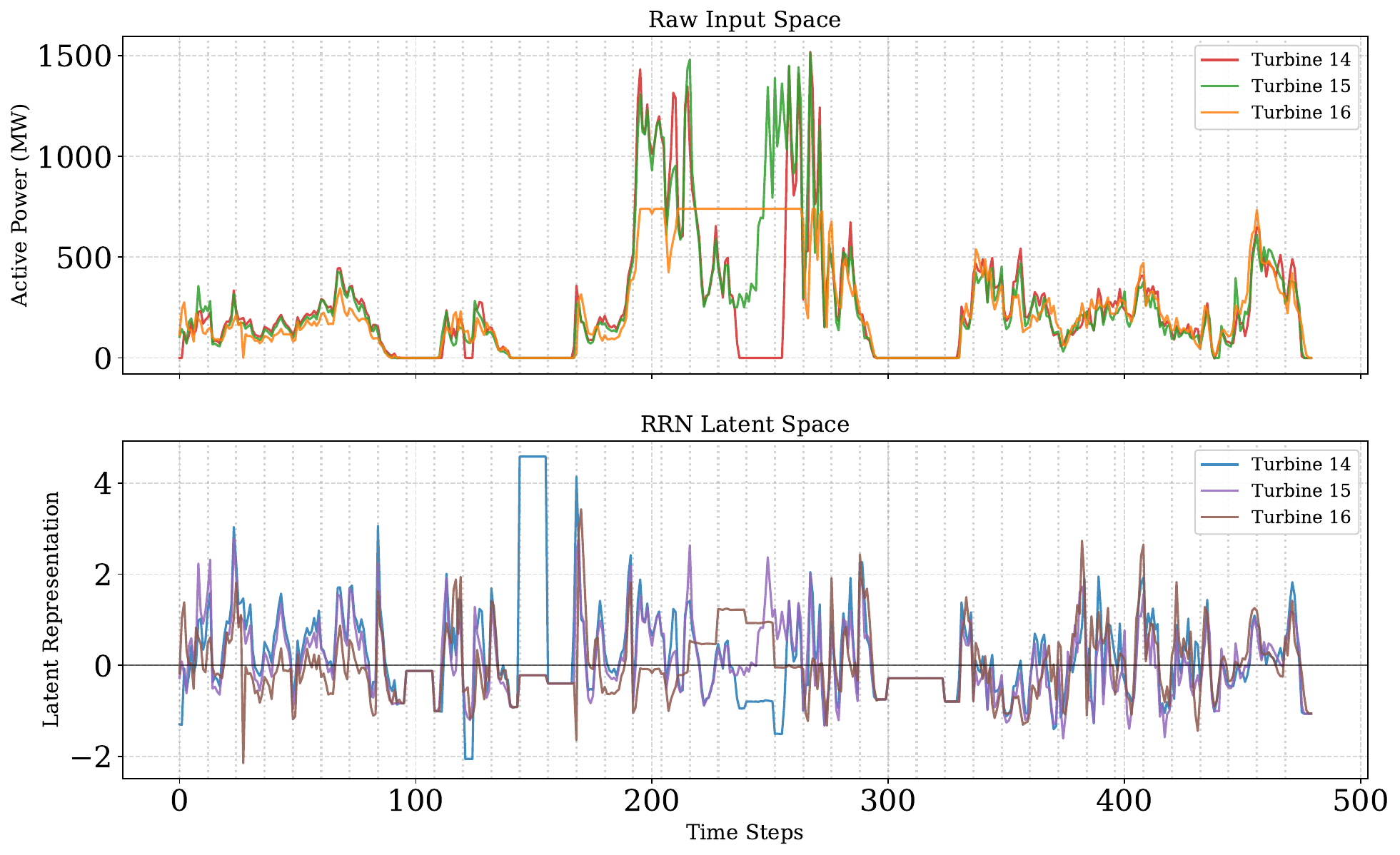}
    \caption{Empirical observation of Spatio-Temporal distribution stabilization on the SDWPF benchmark. The top panel illustrates raw active power trajectories for three geographically neighboring turbines (Turbines 14, 15, and 16), exhibiting significant concept drift, high volatility, and unscaled hardware anomalies (e.g., sensor clipping at $t \in [180, 260]$). The bottom panel displays their corresponding representations in the RRN latent space. Bounded Scale Normalization bounds macro-level non-stationarity into a stable envelope, while the Lipschitz-constrained GNN preserves the micro-spatial heterogeneity (the non-collapsing phase differences among the three neighbors).}
    \label{fig:empirical_stabilization}
\end{figure}

\noindent\textbf{Empirical Proof of Distribution Stabilization} \quad To observe whether the latent space achieves stabilization, we extract a highly volatile temporal window from the SDWPF dataset. Figure~\ref{fig:empirical_stabilization} visualizes the trajectories of three geographically neighboring turbines. In the raw input space, the system experiences significant concept drift—exhibiting large generation peaks—along with real-world sensor anomalies, such as value clipping and curtailment. Directly learning on such non-stationary distributions often destabilizes standard forecasting models. Mapping these signals through our RRN framework yields a stabilized latent space. The Bounded Scale Normalization (BSN) successfully neutralizes the systemic drift, confining the highly volatile sequence to a strictly bounded envelope around zero. More importantly, unlike naive temporal normalizers that would incorrectly homogenize similar variance profiles, the Lipschitz-constrained GNN module ensures that these three adjacent nodes do not collapse into a singular sequence. Their relative phase differences and localized dynamic characteristics are preserved. This visually confirms that RRN mitigates macroscopic distribution shift without destroying microscopic spatial heterogeneity.

\begin{figure*}[htbp]
    \centering
    \includegraphics[width=\textwidth,keepaspectratio]{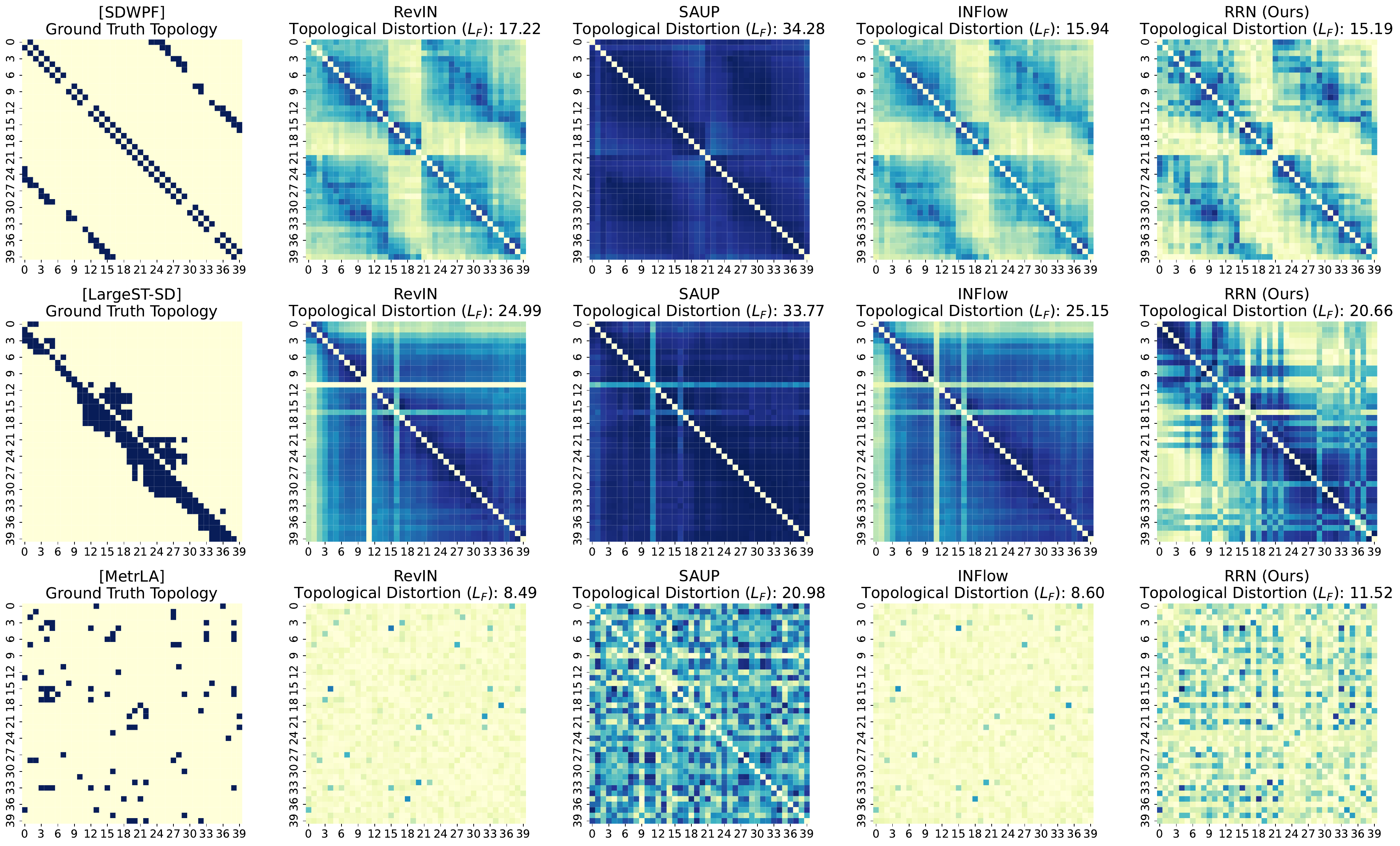}
    \caption{Spatial topology preservation analysis across SDWPF, LargeST-SD, and MetrLA benchmarks. The heatmaps visualize the spatial correlation matrices of latent representations extracted after different normalization modules. The Frobenius norm ($L_F$) quantifies the topological distortion relative to the physical Ground Truth adjacency matrix. RRN yields lower $L_F$, avoiding the structural homogenization observed in conventional temporal normalizers.}
    \label{fig:topology_preservation}
\end{figure*}

\noindent\textbf{Spatial Topology Preservation} \quad A potential limitation of applying pure temporal normalizations to graph-structured data is the disruption of spatial heterogeneity. Standardizing independent temporal sequences can force distinct geographic nodes into identical variance scales, partially obscuring the underlying physical graph topology. To evaluate our framework's performance in addressing this, we compute the structural correlation matrices of the latent features extracted after various normalization modules. The topological distortion is explicitly quantified using the Frobenius norm between the empirical correlation matrix and the ground-truth normalized adjacency matrix: $L_F = \|\text{Corr}(Z) - \tilde{A}\|_F$.

As illustrated in Figure~\ref{fig:topology_preservation}, conventional temporal normalizers (e.g., RevIN, INFlow) enforce unconstrained variance scaling, which tends to homogenize distinct spatial nodes. This results in structural blurring, higher $L_F$ distortion errors on complex networks, and the potential generation of spurious correlations. SAUP attempts spatial alignment but introduces block-wise artifacts that may not capture sparse connectivity well (e.g., $L_F=34.28$ on SDWPF and $L_F=20.98$ on MetrLA). In contrast, RRN utilizes Bounded Scale Normalization coupled with Lipschitz-constrained graph convolutions to adaptively bound variance shifts. By operating symmetrically on the graph structure, RRN mitigates temporal non-stationarity while preserving the intrinsic spatial correlation manifold. As shown in Figure~\ref{fig:topology_preservation}, RRN yields significantly lower Frobenius distortion on large-scale benchmarks such as SDWPF ($L_F=15.19$ vs. $17.22$ for RevIN and $15.94$ for INFlow) and LargeST-SD ($L_F=20.66$ vs. $24.99$ for RevIN and $25.15$ for INFlow), indicating that the framework resolves distribution shifts without severely compromising the graph's structural integrity.
\begin{figure}[t]
    \centering
    \includegraphics[width=\columnwidth,keepaspectratio]{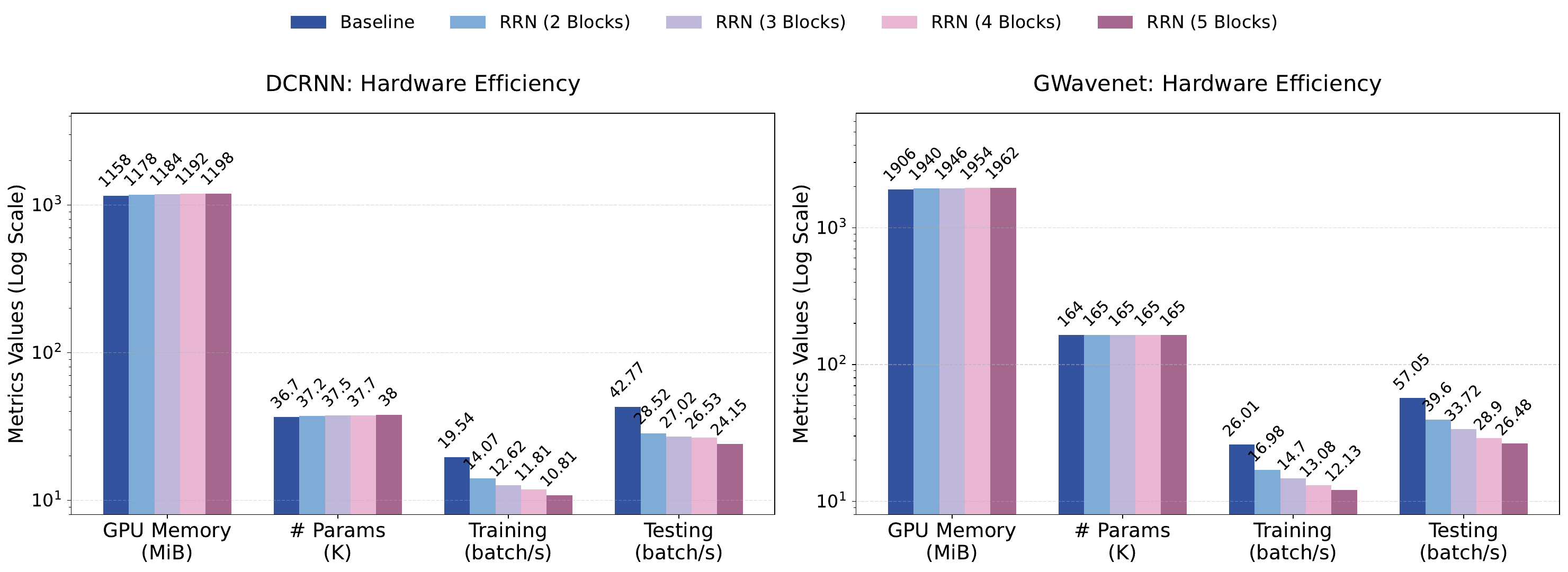}
    \caption{Hardware efficiency comparison between baseline and RRN models (2--5 blocks) across GPU memory, parameter counts, and throughput for DCRNN and Graph WaveNet}
    \label{fig:performance_analysis}
\end{figure}

\begin{figure}[t]
    \centering
    \includegraphics[width=\columnwidth,keepaspectratio]{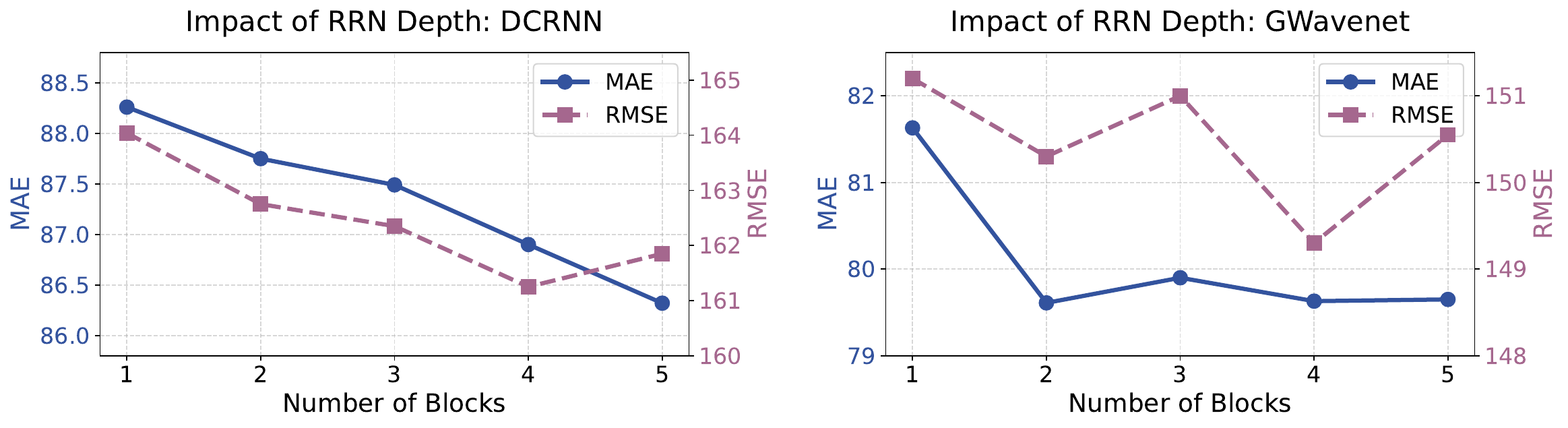}
    \caption{Impact of the number of RRN residual blocks on forecasting accuracy, illustrated through MAE and RMSE metrics for DCRNN and Graph WaveNet architectures}
    \label{fig:Impact_of_RRN_Depth}
\end{figure}

\begin{figure}[t]
    \centering
    \includegraphics[width=\columnwidth,keepaspectratio]{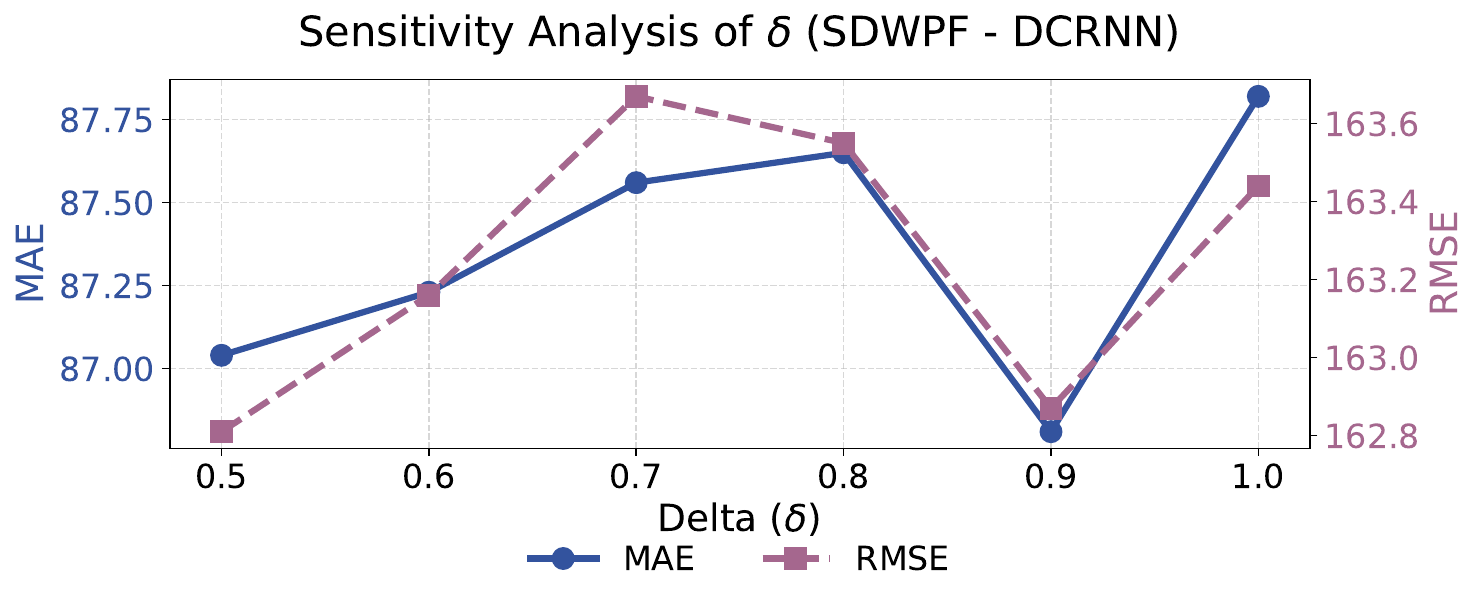}
    \caption{Sensitivity analysis of the Bounded Scale Normalization threshold $\delta$ on the SDWPF dataset using the DCRNN backbone. The performance remains largely stable across different values, indicating the robustness of the bounding mechanism.}
    \label{fig:delta_sensitivity}
\end{figure}

\noindent\textbf{Sensitivity of BSN Threshold ($\delta$)} \quad In our Bounded Scale Normalization (BSN) module, the hyperparameter $\delta$ acts as a lower bound for the variance denominator, which is mathematically crucial for preventing Jacobian singularities and ensuring Lipschitz continuity. To evaluate the framework's sensitivity to this parameter, we tested varying values of $\delta \in [0.5, 1.0]$ on the SDWPF benchmark using DCRNN. As illustrated in Figure~\ref{fig:delta_sensitivity}, the forecasting performance demonstrates stability across the tested range, with MAE fluctuating marginally between 86.81 and 87.82. The optimal performance is achieved around $\delta = 0.9$. This stability empirically confirms that while $\delta$ fulfills its theoretical role of bounding the scale for invertibility, the overall framework does not require exhaustive hyperparameter tuning to maintain accuracy, making it practical for real-world deployments.

\begin{table}[htbp]
\centering
\caption{Ablation study of RRN components on the SDWPF dataset. We evaluate the impact of removing Bounded Scale Normalization (BSN) and the invertible GNN module (iGNN) across two backbone architectures. Best results are highlighted in bold.}
\label{tab:ablation_wide}
\renewcommand{\arraystretch}{1.2}
\begin{adjustbox}{max width=\columnwidth}
\begin{tabular}{@{}lcccccccc@{}}
\toprule
\multirow{2}{*}{\textbf{Variant}} & \multicolumn{3}{c}{\textbf{DCRNN (MAE)}} & & \multicolumn{3}{c}{\textbf{GWavenet (MAE)}} \\ \cmidrule(lr){2-4} \cmidrule(lr){6-8}
 & \textbf{3 Horizon} & \textbf{6 Horizon} & \textbf{12 Horizon} & & \textbf{3 Horizon} & \textbf{6 Horizon} & \textbf{12 Horizon} \\ \midrule
Baseline (Original) & 62.52 & 89.83 & 126.81 & & 56.03 & 82.14 & 117.24 \\
Only ActNorm & 62.86 & 90.12 & 127.74 & & 56.25 & 81.68 & 116.87 \\
w/o BSN & 61.82 & 88.16 & 125.29 & & 55.72 & 80.91 & 116.58 \\
w/o iGNN & 61.34 & 88.64 & 125.57 & & 55.91 & 80.58 & 116.42 \\
w/o Inverse & 124.91 & 136.11 & 157.16 & & 83.02 & 98.99 & 127.09 \\
\textbf{w/ BSN + iGNN (RRN)} & \textbf{60.89} & \textbf{87.41} & \textbf{124.96} & & \textbf{54.86} & \textbf{79.94} & \textbf{116.08} \\ \bottomrule
\end{tabular}
\end{adjustbox}
\end{table}

\begin{table}[t]
\centering
\makegapedcells
\caption{Performance comparison across different prediction horizons on SDWPF dataset. Bold values indicate better performance in each baseline vs. RRN pair (lower is better).}
\label{tab:prediction_length}
\begin{adjustbox}{max width=\columnwidth}
\begin{tabular}{@{}ccccccccc@{}}
\toprule
\multirow{2}{*}{\textbf{Predict length}} & \multicolumn{2}{c}{\textbf{12}} & \multicolumn{2}{c}{\textbf{24}} & \multicolumn{2}{c}{\textbf{48}} & \multicolumn{2}{c}{\textbf{96}} \\
\cmidrule(lr){2-3} \cmidrule(lr){4-5} \cmidrule(lr){6-7} \cmidrule(lr){8-9}
\textbf{Method Avg.} & MAE & RMSE & MAE & RMSE & MAE & RMSE & MAE & RMSE \\
\midrule
DCRNN & 88.11 & 165.48 & 122.61 & 217.74 & 174.94 & 266.56 & 201.82 & 309.45 \\
DCRNN + RRN & \textbf{87.39} & \textbf{163.03} & \textbf{120.67} & \textbf{214.19} & \textbf{166.92} & \textbf{258.33} & \textbf{194.54} & \textbf{304.41} \\
\midrule
GWaveNet & 87.56 & 162.44 & 121.64 & 213.73 & 170.22 & 258.38 & 199.19 & 305.76 \\
GWaveNet + RRN & \textbf{85.76} & \textbf{159.84} & \textbf{119.83} & \textbf{209.47} & \textbf{165.35} & \textbf{252.11} & \textbf{192.69} & \textbf{301.87} \\
\bottomrule
\end{tabular}
\end{adjustbox}
\end{table}

\noindent\textbf{Impact of Prediction Horizon} \quad To evaluate the model's stability over longer sequences, we compare the performance across extended prediction horizons (up to 96 steps) on the SDWPF dataset, as summarized in Table~\ref{tab:prediction_length}. RRN maintains consistent error reductions over the baselines for longer-term predictions, indicating its effectiveness in capturing extended temporal dependencies without accumulating excessive error.\\

\noindent\textbf{Performance Analysis} \quad We evaluate the computational efficiency of RRN across varying depths, as shown in Figure~\ref{fig:performance_analysis}. The framework demonstrates high memory efficiency, maintaining stable GPU usage and constant parameter counts regardless of the number of stacked blocks. However, inference throughput decreases linearly as depth increases. This latency is primarily attributed to the iterative fixed-point calculation required for the inverse transformation, highlighting a trade-off between structural depth and computational speed.\\

\noindent\textbf{Impact of Residual Blocks} \quad We further investigate the relationship between RRN depth and forecasting accuracy, as illustrated in Figure~\ref{fig:Impact_of_RRN_Depth}. Increasing the number of residual blocks generally reduces prediction error, enhancing the model's ability to map complex distributions. However, we observe diminishing marginal returns with greater depth. Our results indicate that a configuration of 2 to 3 blocks achieves the optimal balance between model expressivity and operational efficiency.\\

\noindent\textbf{Ablation Study} \quad We validate the contribution of each localized component within RRN using the gap-filled SDWPF dataset, as detailed in Table~\ref{tab:ablation_wide}. The results demonstrate that removing either Bounded Scale Normalization (BSN) or the invertible GNN module (iGNN) leads to performance degradation. Specifically, BSN is helpful for mitigating non-stationary variance trajectories along the temporal dimension, while iGNN is critical for capturing spatial topological heterogeneity. The complete RRN framework consistently yields lower errors, confirming that the synergy between localized temporal normalization and spatial structure preservation is beneficial for robust forecasting.

\subsection{Reversibility and Sensitivity Analysis}
In this section, we empirically validate the invertibility of our proposed framework and analyze its sensitivity to the Lipschitz constraints. 

\begin{table}[htbp]
\centering
\setlength{\belowcaptionskip}{10pt}
\caption{Sensitivity analysis and reversibility test of RRN. We fix the Bounded Scale Normalization scaling $\alpha = 0.9$ and vary the Frobenius norm constraint $c$.}
\label{tab:sensitivity_gnn_norm}
\renewcommand{\arraystretch}{1.3} 
\setlength{\tabcolsep}{3pt} 
\begin{adjustbox}{max width=\columnwidth} 
\begin{tabular}{@{}c|c|cccccc@{}}
\toprule
\multicolumn{2}{c|}{\multirow{2}{*}{\makecell{\textbf{Config} \\ ($\alpha = 0.9$)}}} & \multicolumn{6}{c}{\textbf{GNN Frobenius Norm ($c$)}} \\
\cmidrule(l){3-8}
\multicolumn{2}{c|}{} & \textbf{0.8} & \textbf{0.9} & \textbf{1.0} & \textbf{5} & \textbf{10} & \textbf{15} \\ \midrule
\multirow{3}{*}{\rotatebox[origin=c]{90}{\textbf{MAE}}} 
& H=3 & 60.45 & 61.27 & 220.04 & 1815.85 & $7.99\!\times\!10^6$ & $9.70\!\times\!10^5$ \\
& H=6 & 86.62 & 86.41 & 464.58 & 1836.87 & $7.63\!\times\!10^6$ & $1.01\!\times\!10^6$ \\
& H=12 & 122.70 & 122.48 & 308.33 & 1836.38 & $7.63\!\times\!10^6$ & $9.14\!\times\!10^5$ \\ \midrule \midrule
\multirow{5}{*}{\rotatebox[origin=c]{90}{\makecell{\textbf{Recon.}\\\textbf{Error}}}} 
& Iter. & \multicolumn{6}{c}{\textbf{Reconstruction Error}} \\ \cmidrule(l){2-8}
& 5 & $3.78\!\times\!10^{-3}$ & $2.46\!\times\!10^{-3}$ & $3.15\!\times\!10^{-2}$ & $8.89\!\times\!10^{-2}$ & $2.58\!\times\!10^{0}$ & $1.30\!\times\!10^{1}$ \\
& 10 & $2.14\!\times\!10^{-3}$ & $1.18\!\times\!10^{-3}$ & $3.16\!\times\!10^{-2}$ & $8.44\!\times\!10^{-2}$ & $1.06\!\times\!10^{1}$ & $3.36\!\times\!10^{2}$ \\
& 20 & $1.23\!\times\!10^{-3}$ & $4.89\!\times\!10^{-4}$ & $3.25\!\times\!10^{-2}$ & $1.15\!\times\!10^{-1}$ & $7.36\!\times\!10^{1}$ & $1.49\!\times\!10^{5}$ \\
& 50 & $2.94\!\times\!10^{-4}$ & $5.92\!\times\!10^{-5}$ & $3.30\!\times\!10^{-2}$ & $2.91\!\times\!10^{-1}$ & $1.07\!\times\!10^{5}$ & $7.18\!\times\!10^{13}$ \\ \bottomrule
\end{tabular}
\end{adjustbox}
\end{table}

\begin{table}[htbp]
\centering
\setlength{\belowcaptionskip}{10pt}
\caption{Sensitivity analysis and reversibility test of RRN. We fix the GNN Frobenius norm constraint $c = 0.9$ and vary the Bounded Scale Normalization scaling $\alpha$.}
\label{tab:sensitivity_cn_scaling}
\renewcommand{\arraystretch}{1.3} 
\setlength{\tabcolsep}{3pt} 
\begin{adjustbox}{max width=\columnwidth} 
\begin{tabular}{@{}c|c|cccccc@{}}
\toprule
\multicolumn{2}{c|}{\multirow{2}{*}{\makecell{\textbf{Config} \\ ($c = 0.9$)}}} & \multicolumn{6}{c}{\textbf{BSN Scaling ($\alpha$)}} \\
\cmidrule(l){3-8}
\multicolumn{2}{c|}{} & \textbf{0.8} & \textbf{0.9} & \textbf{1.0} & \textbf{5} & \textbf{10} & \textbf{15} \\ \midrule
\multirow{3}{*}{\rotatebox[origin=c]{90}{\textbf{MAE}}} 
& H=3 & 62.64 & 61.33 & 258.54 & 72.97 & $1.27\!\times\!10^7$ & $1.57\!\times\!10^6$ \\
& H=6 & 88.33 & 86.93 & 258.56 & 106.45 & $9.56\!\times\!10^5$ & $1.47\!\times\!10^6$ \\
& H=12 & 124.80 & 123.22 & 258.47 & 134.29 & $9.02\!\times\!10^5$ & $1.14\!\times\!10^6$ \\ \midrule \midrule
\multirow{5}{*}{\rotatebox[origin=c]{90}{\makecell{\textbf{Recon.}\\\textbf{Error}}}} 
& Iter. & \multicolumn{6}{c}{\textbf{Reconstruction Error}} \\ \cmidrule(l){2-8}
& 5 & $1.72\!\times\!10^{-3}$ & $2.52\!\times\!10^{-3}$ & $1.37\!\times\!10^{-1}$ & $2.01\!\times\!10^{-1}$ & $3.63\!\times\!10^1$ & $4.72\!\times\!10^0$ \\
& 10 & $3.42\!\times\!10^{-4}$ & $9.39\!\times\!10^{-4}$ & $1.64\!\times\!10^{-1}$ & $2.75\!\times\!10^{-1}$ & $9.06\!\times\!10^3$ & $6.04\!\times\!10^1$ \\
& 20 & $4.59\!\times\!10^{-5}$ & $3.17\!\times\!10^{-4}$ & $1.70\!\times\!10^{-1}$ & $3.11\!\times\!10^{-1}$ & $7.39\!\times\!10^7$ & $7.04\!\times\!10^4$ \\
& 50 & $7.96\!\times\!10^{-7}$ & $3.43\!\times\!10^{-5}$ & $1.73\!\times\!10^{-1}$ & $3.26\!\times\!10^{-1}$ & $5.92\!\times\!10^{20}$ & $2.87\!\times\!10^{15}$ \\ \bottomrule
\end{tabular}
\end{adjustbox}
\end{table}

We validate the invertibility condition from Lemma~\ref{lemma:invertible_residual} by analyzing forecasting accuracy (MAE) and reconstruction error. As shown in Table~\ref{tab:sensitivity_gnn_norm} and Table~\ref{tab:sensitivity_cn_scaling}, increasing the GNN constraint $c$ or Bounded Scale Normalization scaling $\alpha$ gradually violates the contraction mapping condition. When parameters remain strictly below 1, the model maintains low forecasting error and negligible reconstruction error ($10^{-3}$--$10^{-4}$). However, as $c$ or $\alpha$ reaches or exceeds 1, forecasting precision deteriorates significantly, and the reconstruction error explodes ($>10^6$) as the fixed-point iteration diverges. These results empirically confirm that enforcing the Lipschitz constant ($<1$) is essential for both training stability and the mathematical validity of the inverse transformation.

\section{Conclusion}
This study addresses the critical challenge of distribution shift in spatio-temporal forecasting, characterized by simultaneous temporal non-stationarity and spatial heterogeneity. To mitigate this, we propose Reversible Residual Normalization (RRN), a novel framework that integrates Lipschitz-continuous Bounded Scale Normalization with spectral-constrained graph convolutions. By enforcing invertibility, RRN projects complex spatio-temporal data into a stationary latent space while preserving topological dependencies. Extensive experiments demonstrate that our model-agnostic approach consistently enhances the robustness and generalization of deep forecasting models by explicitly accounting for coupled spatial and temporal distribution shifts.

\bibliographystyle{IEEEtran}
\bibliography{reference}

\end{document}